# Predicting Continuous Locomotion Modes via Multidimensional Feature Learning from sEMG

Peiwen Fu[1], Wenjuan Zhong[1], Yuyang Zhang[1], Wenxuan Xiong[1], Yuzhou Lin[1], Yanlong Tai[2], Lin Meng[3] and Mingming Zhang[1], *Senior Member, IEEE*

**Abstract**—Walking-assistive devices require adaptive control methods to ensure smooth transitions between various modes of locomotion. For this purpose, detecting human locomotion modes (e.g., level walking or stair ascent) in advance is crucial for improving the intelligence and transparency of such robotic systems. This study proposes Deep-STF, a unified end-to-end deep learning model designed for integrated feature extraction in spatial, temporal, and frequency dimensions from surface electromyography (sEMG) signals. Our model enables accurate and robust continuous prediction of nine locomotion modes and 15 transitions at varying prediction time intervals, ranging from 100 to 500 ms. In addition, we introduced the concept of 'stable prediction time' as a distinct metric to quantify prediction efficiency. This term refers to the duration during which consistent and accurate predictions of mode transitions are made, measured from the time of the fifth correct prediction to the occurrence of the critical event leading to the task transition. This distinction between stable prediction time and prediction time is vital as it underscores our focus on the precision and reliability of mode transition predictions. Experimental results showcased Deep-STP's cutting-edge prediction performance across diverse locomotion modes and transitions, relying solely on sEMG data. When forecasting 100ms ahead, Deep-STF surpassed CNN and other machine learning techniques, achieving an outstanding average prediction accuracy of 96.48%. Even with an extended 500ms prediction horizon, accuracy only marginally decreased to 93.00%. The averaged stable prediction times for detecting next upcoming transitions spanned from 28.15 to 372.21 ms across the 100-500 ms time advances.

## I. INTRODUCTION

EXOSKELETON technology has emerged as a promising solution to promote lower-limb motor function by providing support and assistance to individuals with mobility impairments [1-6]. However, despite its potential, exoskeletons face challenges when it comes to adapting to the requirements and motions of human users [7-9]. Poor adjustments can result in discomfort, inefficient gait patterns, and even safety concerns, thus impeding clinical acceptance [10-13]. For instance, users may experience frustration and instability when the reactions of the exoskeleton do not align with their movements, especially during sudden changes. A straightforward solution to address this involves integrating gait mode recognition and prediction mechanisms. Robust and accurate prediction of locomotion intentions can significantly enhance exoskeleton adaptability, leading to smoother and more intuitive interactions between users and assistive devices.

Continuous prediction of locomotion intentions presents a significant challenge stemming from two critical aspects. Firstly, the real-time implementation of the lower-limb exoskeleton necessitates the continuous prediction of both steady and transitional states. Steady states encompass periods of consistent, unchanging locomotion patterns, characterized by relatively stable, predefined gait modes like walking, jogging, or standing still. In contrast, transitional states capture the moments when an individual is transitioning from one locomotion mode to another, marked by rapid and complex shifts in movement patterns. Predicting transitional states is notably more challenging, as many studies have consistently reported lower accuracy levels in comparison to steady states [14-18]. The underlying reasons for this discrepancy can be attributed to the intricacy, temporal variability, limited training data, and the abrupt changes inherent to these locomotion modes [19]. The second aspect of this challenge revolves around locomotion intention prediction, demanding the ability to anticipate a locomotion intention in advance based on current information [20]. This challenge is further compounded when extended prediction times are required [21, 22]. For instance, the increase of prediction time from 50 to 200 ms in [23] significantly amplified errors, with the root mean square error increasing from 0.68 to 4.62 degrees for knee flexion angles.

In summary, these dual challenges stand as pivotal hurdles that must be overcome to facilitate the adaptation of rehabilitation or assistive devices, ultimately ensuring their

The work was supported in part by the National Key Research and Development Program of China under Grant 2022YFF1202500 and Grant 2022YFF1202502, in part by the Natural Science Foundation of Shenzhen under Grant JCYJ20210324104203010, in part by Shenzhen Key Laboratory of Smart Healthcare Engineering under Grant ZDSYS20200811144003009, in part by the Research Foundation of Guangdong Province under Grant 2019ZT08Y191. (Peiwen Fu, Wenjuan Zhong and Yuyang Zhang contributed equally to this work.) (*Corresponding authors: Mingming Zhang.*)

1 Peiwen Fu, Wenjuan Zhong, Yuyang Zhang, Wenxuan Xiong Yuzhou Lin, and Mingming Zhang are with the Shenzhen Key Laboratory of Smart Healthcare Engineering, Department of Biomedical Engineering, College of Engineering, Southern University of Science and Technology, Shenzhen 518055, China. (e-mail: zhangmm@sustech.edu.cn.)
2 Yanlong Tai is with Shenzhen Institutes of Advanced Technology, CAS, Shenzhen, 518055, China.
3 Lin Meng is with Academy of Medical Engineering and Translational Medicine, Tianjin University, Tianjin, 300192, China.

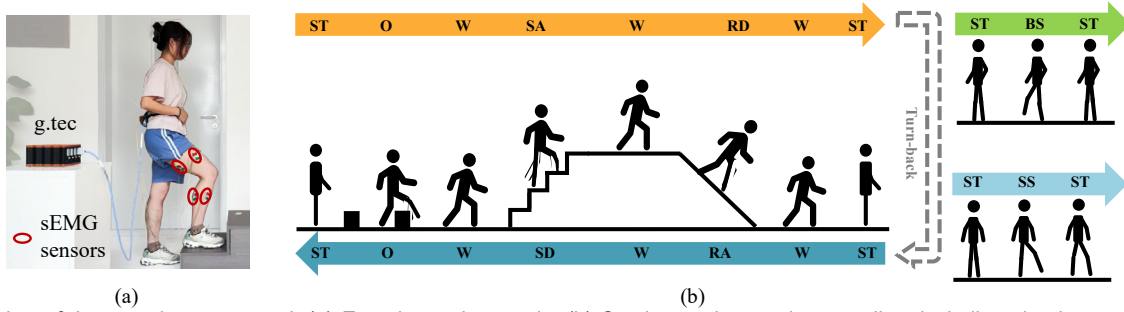

Fig. 1. Overview of the experiment protocol. (a) Experimental scenario. (b) Continuous locomotion paradigm including nine locomotion modes: Standing (ST), Obstacle (O), Level Walking (W), Stair Ascent (SA), Ramp Descent(RD), Ramp Ascent (RA), Stair Descent (SD), Back-Stepping (BS), and Side-Stepping mode (SS); There exists 15 transitional states among these locomotions: ST→O, O→W, W→SA, SA→W, W→RD, RD→W, ST→W, W→RA, RA→W, W→SD, SD→W, W→O, O→ST, ST→BS, and ST→SS.

capacity to effectively cater to the changing requirements and evolving movements of their users. To address them, current solutions can be categorized as unsupervised or supervised, based on the type of learning algorithms employed. For either method, it is commonly to see that model prediction capabilities can be further enhanced by employing more classifiers or augmenting more sensors.

In terms of unsupervised learning, Afzal et al. [24] explored muscle synergy extracted from surface electromyography (sEMG) signals, which served as neural features for an unsupervised machine-learning algorithm aimed at predicting locomotion modes. However, their findings revealed that the extracted muscle synergy had relatively high classification errors in identifying transitions. To overcome this limitation, Liu et al. [25] proposed a different approach by incorporating the information of linear acceleration and angular velocity from inertial measurement unit (IMU) sensors alongside sEMG signals. They trained six classifiers specifically designed for transition identification, significantly enhancing the potential of using muscle synergy for detecting locomotion intentions. In addition, traditional supervised machine learning methods are frequently employed to enhance locomotion prediction capabilities. For instance, Peng et al. [26] proposed a multi-level classifier fusion strategy, utilizing random forest models in conjunction with both sEMG signals and IMUs to improve performance in complex operational environments. When it comes to data fusion, integrating other mechanical sensors like load cells [27] and capacitive sensors [28] can enhance the prediction of locomotion patterns.

While several studies have demonstrated the enhanced effectiveness of integrating sEMG with additional sensors for improved locomotion intention recognition, Meng et al. [29] further emphasized the significance of sEMG by specifically investigating its role in data fusion when combined with IMU for this task. In addition, Zhang et al. [30] pointed out the importance of sensor selection for the control of powered artificial legs and demonstrated the superiority of sEMG signals. These studies underscored the substantial potential of sEMG, raising an intriguing question: Can this challenging task be effectively addressed by relying solely on sEMG? Continuous prediction of locomotion intentions using traditional machine learning models or even conventional deep learning methods based solely on sEMG is often challenged by their limited capacity to capture subtle variations in muscle activation patterns [31]. This limitation constrains the number of detectable locomotion modes and the prediction time. We argue that the untapped potential of sEMG can be harnessed by designing advanced deep-learning models capable of extracting robust features from both temporal and spatial domains. The study by Fu et al. [32] serves as a compelling example, wherein they incorporated spatial and temporal convolutions for joint angle analysis during continuous gait tasks, achieving remarkable results and providing valuable insights.

The primary objective of this study was to enhance the continuous prediction of locomotion intentions exclusively using sEMG signals. To achieve this, we designed a continuous locomotion paradigm that encompassed a wider range of locomotion modes and transitions (nine locomotion modes and 15 transitional states), thereby enhancing its representational diversity. Subsequently, we introduced Deep-STF, an innovative multidimensional feature learning network with the capacity to extract sEMG features from spatial, temporal, and frequency domains, along with an adaptive voting strategy. To assess the prediction capabilities, we introduced two critical metrics: transitional-state accuracy and stable prediction time. The main contributions of this study are summarized as follows: (1) We propose a unified end-to-end framework that incorporates feature extraction in spatial, temporal, and frequency dimensions from sEMG signals for continuous mode prediction. (2) Human subjects experiments validate the feasibility of our approach, achieving a remarkable prediction accuracy of 96.48% and reporting the best averaged stable prediction time for detecting upcoming transitions at 372.21ms.

## II. METHODOLOGY

### A. Participants and Experiment Protocol

Eight subjects without any musculoskeletal disorders or recent lower-extremity injuries (5F/3M, height: 164.5 ± 10.2 cm, weight: 58.5 ± 13.1 kg, age: 19 ± 1.5 years) were recruited to take part in the experiments. Fig. 1(a) presents the experimental scenarios. The study was approved by the Southern University of Science and Technology, Human Participants Ethics Committee (20190004), and consents were obtained from the subjects.

As illustrated in Fig. 1(b), a stairs-platform-ramp module with stairs of 18cm in height, a platform with 1.8 m, and an inclination angle of 15° was used in the experiments on continuous gait tasks. Two square-shaped obstacles of 20 cm in height were placed on the left side of the module at a suitable distance. In each trial, subjects started by standing still, then walked from one side of the module to the other side, and then reversed back to the origin.

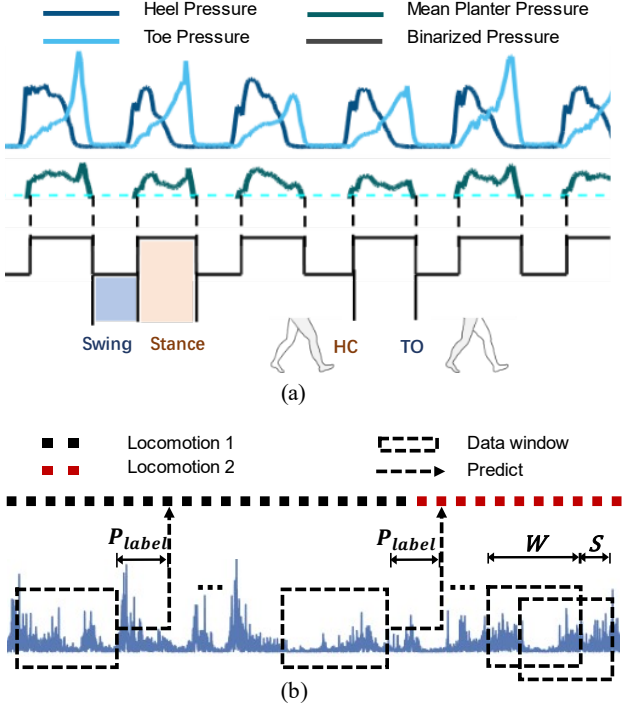

Fig. 2. Plantar pressure data processing and data labelling assignment. (a) Plantar pressure data processing to determine heel-contact (HC) and toe-off (TO) of the leading leg. (b) sEMG segmentation and class labeling. The sliding window length is represented as $W$, and the stride is represented as $S$. $P_{label}$ is the prediction time for class labeling.

There were seven locomotion and 12 locomotion transitions. Each subject started from standing mode (ST), standing mode to obstacle mode (ST→O), obstacle mode (O), obstacle mode to level walking mode (O→W), level walking mode (W), level walking mode to stair ascent mode (W→SA), stair ascent mode (SA), stair ascent mode to level walking mode (SA→W), level walking mode, level walking mode to ramp descent mode (W→RD), ramp descent mode (RD), ramp descent mode to level walking mode (RD→W), level walking mode, level walking to standing mode. After a turn-back step with no data collection, standing to level walking (ST→W), level walking, level walking to ramp ascent (W→RA), ramp ascent (RA), ramp ascent to level walking (RA→W), level walking to stair descent (W→SD), stair descent (SD), stair descent to level walking (SD→W), level walking, level walking to obstacle (W→O), obstacle and finally standing (O→ST).

In addition, each subject performed more motions, including (1) stand-to-back stepping, where they first stood still and then took two steps backward, and (2) stand-to-side stepping, where they first stood still and then took two steps to the side. These motions encompass three locomotion modes: standing mode, back-stepping mode (BS), and side-stepping mode (SS). Furthermore, there were two locomotion transitions between them: standing mode to back-stepping mode (ST→BS) and standing to side-stepping mode (ST→SS). Each task was repeated 50 times, with adequate rest periods provided after every five trials to mitigate muscle fatigue.

Before the formal experiment, all participants underwent preliminary practice to ensure consistent step counts among subjects. In each gait transition, the right leg equipped with electrodes was designated as the leading leg. Throughout the entire experiment, subjects were encouraged to walk at their self-selected, comfortable speed and to make efforts to maintain this speed as consistently as possible.

### B. Data Collection and Preprocessing

The sEMG data were recorded from eight muscles on the right lower limb with a bio-signal amplifier (g.HIamp-Research, Austria). The selected muscles were biceps femoris (BF), semitendinosus (SM), medial gastrocnemius (MG), lateral gastrocnemius (LG), vastus medialis (VM), vastus lateralis (VL), rectus femoris (RF), and tibialis anterior (TA). Bipolar active-type Ag-AgCl electrodes were placed according to SENIAM recommendations [33]. The sEMG data were recorded at a sample rate of 1200 Hz. A 50Hz notch filter was applied to reduce powerline interference. A two-channel thin-film plantar pressure sensor was placed under the heel and toe of the right foot to record heel and toe pressures, with a sample rate of 40 Hz.

The EMG signals were preprocessed using custom analysis code in MATLAB software. An $8^{th}$ order Butterworth filter with a 20-500 Hz cutoff frequency was applied to raw sEMG signals. All negative values after filtering were rectified by taking their absolute values. Then, the sEMG data were normalized across all channels using the mean and standard deviation of each channel. Furthermore, plantar pressure data were up-sampled to 1200Hz using linear interpolation to align the sampling rate with that of sEMG. Because of sensor noise, the pressure signals from the heel and toe were normalized and then averaged to create the mean plantar pressure signal, as presented in Fig. 2(a). Then, this signal underwent a binarization process to determine the critical gait events, heel-contact (HC) and toe-off (TO) of the leading leg. These events served as key points in determining mode transitions between different locomotion modes.

In our continuous locomotion scenarios, we have listed the critical events that are used to differentiate between various locomotion modes in Table I. Continuous motions encompass both transition-state and steady-state conditions. Transition-state conditions occur when subjects exhibit intentions and behaviors related to mode transitions. Without standardized criteria, these conditions are often defined as one or half gait cycles [14, 15, 24, 25] or specific time durations during which mode transition points occur [26]. In line with the research

TRANSITION BETWEEN LOCOMOTION MODES. "HC": HEEL CONTACT; "TO": TOE OFF; "LEAD": LEADING LEG IN CERTAIN LOCOMOTION.

| Mode Transitions | Mode Transition Points | Mode Transitions | Mode Transition Points |
|---|---|---|---|
| W to O | Lead HC | O to W | Lead HC |
| W to SA | Lead HC | SA to W | Lead HC |
| W to SD | Lead HC | SD to W | Lead HC |
| W to RA | Lead HC | RA to W | Lead HC |
| W to RD | Lead HC | RD to W | Lead HC |
| W to ST | Lead HC | ST to W | Lead TO |
| ST to SS | Lead TO | ST to BS | Lead TO |
| ST to O | Lead TO | | |

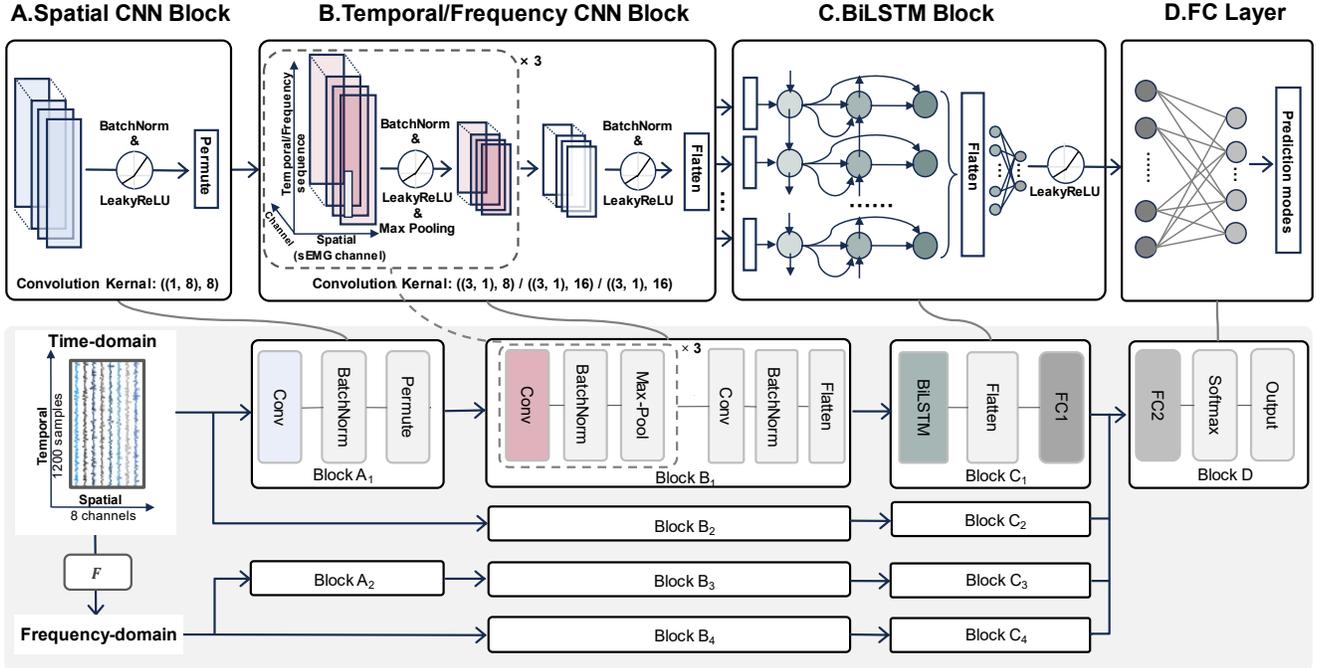

Fig. 3. Detailed illustration of the proposed Deep-Spatial-Temporal-Frequency (Deep-STF) model for continuous locomotion mode prediction. Information from time-series data from sEMG sensors is captured continuously with a defined input window and also further transformed into a frequency domain window. Four batches extract relevant features and map the raw sensor data to nine locomotion modes.

findings from [25], in this study, the transitional state was defined as the time period of 0-500ms before the occurrence of mode transition points. On the other hand, the steady state referred to time intervals characterized by the absence of transition intentions.

### C. Data Label Assignment

Fig. 2(b) illustrates the data labeling strategy. Data were segmented using a sliding window approach with a window length, denoted as $W$, of 1200 time samples and a stride, presented as $S$, of 60 time samples. Each sEMG data window was labeled based on the locomotion that happened in the subsequent $P_{label}$ time samples. $P_{label}$, referring to the prediction time for labeling, signifies that each data window is labeled according to the locomotion that will occur after a time period of $P_{label}$.

### D. Network Architecture

The Deep-STF model, a part of our proposed framework, facilitates the extraction of sEMG features from the spatial, temporal, and frequency domains. As presented in Fig.3, Deep-STF is a Pseudo-Siamese network consisting of four branches. The first branch consists of a spatial Convolutional Neural Network (CNN) block and a temporal CNN block in series with a Bidirectional Long Short-Term Memory (BiLSTM) block, extracting features in spatial-temporal domain. The second branch consists of a temporal CNN block and a BiLSTM block in series for temporal feature extraction. The third branch is composed of a spatial CNN block and a frequency CNN block and a BiLSTM block in series to extract spatial-frequency features in the fast Fourier transform of sEMG signals. The input of the fourth branch is also the fast Fourier transform of sEMG signals, and it consists of a frequency CNN block and a BiLSTM block in series for feature extraction in the frequency domain. All the high-level features extracted from each branch were integrated within the final fully connected layer to calculate the output classification probability using a softmax layer. Furthermore, we employed commonly used techniques including batch normalization (BatchNorm), Leaky Rectified Linear Unit (Leaky-Relu) activation function, and max pooling layers. More details can be found in Table II.

In the spatial CNN block, we fitted eight convolution kernels of size (1, 8). Input shape and output shape were (1, 1200, 8), then we permuted the kernel dimension and channel dimension. This allowed us to extract spatial features without changing the shape. In the temporal CNN block, we performed four CNN layers in sequence. Each of the four CNN layers used convolution kernels of size (3, 1). The output of the eight-channel feature maps transformed the temporal/frequency domain sEMG signal into (16, 16, 8) before flattening, resulting in a final output feature of (16, 128). The structure of the frequency CNN block mirrored that of the temporal CNN block, with the frequency CNN block focusing on extracting frequency domain features. In the BiLSTM block, the input data was first fed into the forward Long Short-Term Memory (LSTM) layer, and then reversed and passed through the other backward LSTM layer. Each LSTM layer contained 128 LSTM cells, each with 64 hidden units. The BiLSTM layer extracted features after the temporal/frequency CNN blocks.

Following the Deep-STF model, we implemented an adaptive voting layer to further improve the stability of the model. The adaptive voting layer essentially comprises a two-layer fully connected neural network. Its input comprises the five time-step output classification probabilities generated by the Deep-STF model. These probabilities are concatenated and flattened into a single sequence of size (1, 45). The final predicted class is determined by aggregating the votes across

| Modules | Layers | Parameters | Output |
|---|---|---|---|
| Input | - | - | 1×1200×8 |
| Spatial block | Conv | 1×8, 8 | 8×1200×1 |
| | Permute | - | 1×1200×8 |
| | BatchNorm | - | 1×1200×8 |
| | Leaky-Relu | 0.01 | 1×1200×8 |
| Temporal/Frequency block | Conv | 3×1, 8 | 8×1198×8 |
| | BatchNorm | - | 8×1198×8 |
| | Leaky-Relu | 0.01 | 8×1198×8 |
| | Max Pooling | 4×1, stride 4 | 8×299×8 |
| | Conv | 3×1, 16 | 16×297×8 |
| | BatchNorm | - | 16×297×8 |
| | Leaky-Relu | 0.01 | 16×297×8 |
| | Max Pooling | 4×1, stride 4 | 16×74×8 |
| | Conv | 3×1, 16 | 16×72×8 |
| | BatchNorm | - | 16×72×8 |
| | Leaky-Relu | 0.01 | 16×72×8 |
| | Max Pooling | 4×1, stride 4 | 16×18×8 |
| | Conv | 3×1, 16 | 16×16×8 |
| | BatchNorm | - | 16×16×8 |
| | Leaky-Relu | 0.01 | 16×16×8 |
| | Flatten | - | 16×128 |
| BiLSTM | Bi-Lstm | layers 3 | 16×128 |
| | Flatten | - | 1×2048 |
| | Fully-connected 1 | 2048, 64 | 1×64 |
| | Leaky-Relu | 0.01 | 1×64 |
| | Concatenation | - | 1×256 |
| Fully connected layer | Fully-connected 2 | 256, 9 | 1×9 |
| | SoftMax | - | 1×9 |
| Adaptive voting | Concatenation | - | 5×9 |
| | Flatten | - | 1×45 |
| | Fully-connected 3 | 45, 9 | 1×9 |
| | SoftMax | - | 1×9 |

five time steps.

### E. Model Training Strategy

The proposed model was optimized by a combination of adaptive moment estimation (Adam) with betas of (0.9, 0.99), an learning rate reduction on plateau (ReduceLROnPlateau). The initial learning rate for the model was set at 0.001, which was dynamically adjusted by ReduceLROnPlateau based on the training average loss. Early stopping was adopted to avoid overfitting. All models were trained on a single NVIDIA Tesla V100 GPU with a batch size of 256.

In each subject-specific dataset, all trials were randomly shuffled and then divided into five non-overlapping groups for five-fold cross-validation. Within each fold, sEMG trials were partitioned into two training sets and one test set in a ratio of 7:1:2, denoted as TrainSet1, TrainSet2, and TestSet, respectively. The overall training process of the model included two steps. In the first training step, the Deep-STF network was trained based on the pre-processed sEMG signals. In the second training step, the adaptive voting module was trained based on the output results of Deep-STF module.

In the first training step, TrainSet1 was used as the training set, TrainSet2 was used as the validation set and TestSet was used as the test set to train the Deep-STF module. As the main part of the whole model, Deep-STF module was expected to independently complete the prediction task, so the teacher forcing strategy was used. The standard binary cross entropy (BCE) loss was adopted as the loss function for Deep-STF,

$$L(y, \hat{y}) = -\frac{1}{N} \sum_{i=1}^{N} [y_i \log(\hat{y}_i) + (1 - y_i) \log(1 - \hat{y}_i)] \quad (11)$$

where $y_i$ and $\hat{y}_i$ represented the label and prediction for motion $i$.

In the second training step, TrainSet2 was used as the training set, and TestSet was used as the test set to train the adaptive voting module. It is worth noting that since there was no validation set in Step 2, the ReduceLROnPlateau and early stopping strategy were not used. The module optimization also utilized the BCE loss. When the module began, for the data prior to this moment, where there were no earlier time series, we used the label 0 to fill in the gaps.

### F. Performance Evaluation

The performance of the proposed methods is evaluated from two perspectives: prediction effectiveness and prediction accuracy. Although data windows are labeled with a fixed labeled prediction time, the actual prediction effectiveness is comprehensively analyzed. In transitional state, a prediction is deemed stable when the classifier consistently generates five correct predictions for the upcoming switched locomotion mode. This definition of a stable prediction for transitions is designed to mitigate the impact of noise-induced and fluctuation-induced transition determinations. Based on this, the stable prediction time, denotes as $t_{stable}$, is defined as the time elapsed from the time of the fifth correct prediction for transition ($t_d$) to the time of the critical event of task transition ($t_c$), calculated as:

$$P_{stable} = t_c - t_d. \quad (12)$$

When $P_{stable}$ is positive, it indicates that our model could predict the next locomotion switch before gait transition occurs. Conversely, a negative stable prediction time implies that our model recognizes the next locomotion switch after the gait transition. Therefore, we defined a predict rate using the following equation:

$$Predict\ Rate = \frac{N_{ps}}{N_t}, \quad (13)$$

where $N_t$ presents the count of transitions within the continuous locomotion, and $N_{ps}$ stands for the number of predictions with a positive $P_{stable}$.

We calculated prediction accuracy based on both steady state and transitional state, denoted as $Acc_{ss}$ and $Acc_{TS}$. $Acc_{ss}$ measures the percentage of predicted locomotion that aligns with the actual ground truth locomotion modes and is calculated as:

$$Acc_{ss} = \frac{TP_{ss}}{N_{ss}}, \quad (14)$$

where $TP_{ss}$ presents the number of predictions that aligns with the ground truth locomotion modes, and $N_{ss}$ represents the total number of data windows in steady states. In addition, $Acc_{TS}$ is defined as the percentage of stably predicted locomotion that matches the previous locomotion before $t_d$ as well as next switched locomotion after $t_d$, as

$$Acc_{TS} = \frac{\sum_{i=1}^{N_{TS}} \delta(p_{stable}(i),\ p_{pred}(i))}{N_{TS}}, \quad (15)$$

where $\delta(\cdot, \cdot) \in \{0,1\}$ is an equivalence indicator. $N_{TS}$ represents the number of data windows in transitional states.

According to the stable prediction, we defined a vector $p_{stable}$, representing a series of N labels with adjusted stable prediction time in a transitional state. In $p_{stable}$, modes before $t_d$ represent the locomotion before the transition, while elements after $t_d$ correspond to the subsequent switched locomotion. Moreover, a vector $p_{pred}$ illustrates the output locomotion-mode series of the proposed classifier.

### G. Statistical Analysis

To compare the performance of Deep-STF model and the adaptive voting strategy, a one-way analysis of variance (ANOVA) was used. The normality of the classification results were tested using the Shapiro-Wilk test. After ANOVA, we conducted a *post hoc* analysis with a Tukey's multiple comparisons test method to compute the statistical difference across different classifiers. A significance threshold of $p = 0.05$ was employed in all tests.

## III. RESULTS

### A. Accuracy of Continuous Locomotion Prediction

The accuracy of continuous locomotion prediction was evaluated under steady state and mode transitional state. Prediction accuracies with $P_{label}$ of 250 ms are presented in Fig. 4(a), including comparisons with three traditional feature-based methods (i.e., SVM, LDA, and RF) used in previous works [27, 34-36], as well as CNN for [37]. All models were trained in the same way. Results showed that our proposed Deep-STF model yielded the highest average accuracies across eight subjects. With $P_{label}$ of 250 ms, the Deep-STF model achieved prediction accuracies of 95.40±1.62% for steady state and, 93.79±1.04% for transitional state. The overall accuracy across these conditions is 95.10±1.49%. For CNN model in the same condition, the accuracies for steady state and transitional state are 91.14±3.38% and 87.99±3.54%, with an overall accuracy of 90.57±3.38%. Among the five methods, the Deep-STF achieved significantly higher accuracies in all conditions.

In addition, the adaptive voting strategy enhanced the prediction accuracy of the deep learning-based methods. When combining CNN with adaptive voting, a substantial improvement of 2.04%, 4.06%, and 2.40% was observed in steady-state, transition-state, and overall accuracy, respectively, compared to using the CNN model alone. Additionally, the adaptive voting strategy improved the prediction accuracy of the Deep-STF model, resulting in increases of 0.2%, 0.83%, and 0.31% in steady, transitional state, and overall accuracy, respectively. The introduction of the adaptive voting strategy to both CNN and Deep-STF significantly performed better in transitional-state and overall accuracies.

### B. Accuracy of Different Prediction Time for Labelling

We explored how different prediction times for labeling affected prediction accuracy. As $P_{label}$ extended from 100 ms to 500 ms, both CNN and Deep-STF methods exhibited varying declines in prediction accuracy, as depicted in Fig. 5. For the CNN method, with a $P_{label}$ of 100 ms, the steady-state, transition-state, and overall accuracy stood at 92.66%, 89.59%, and 92.11%, respectively. When $P_{label}$ increased to 500 ms, these accuracies reduced slightly to 85.31%, 83.76%, and 85.02%, representing a decrease of 7.34%, 5.83%, and 7.09%. Notably, locomotion prediction using the Deep-STF model consistently delivered significantly higher accuracy compared to CNN (p < 0.01). At $P_{label}$ of 100 ms, the steady-state, transition-state, and overall accuracy were 96.43%, 95.45%, and 96.25%, respectively. With an increased $P_{label}$ of 500 ms, these numbers decreased marginally to 92.54%, 90.47%, and 92,13%, with a reduction of 3.90%, 4.97%, and 4.12%. The proposed Deep-STF method demonstrated a lesser decline in accuracy with increasing $P_{label}$.

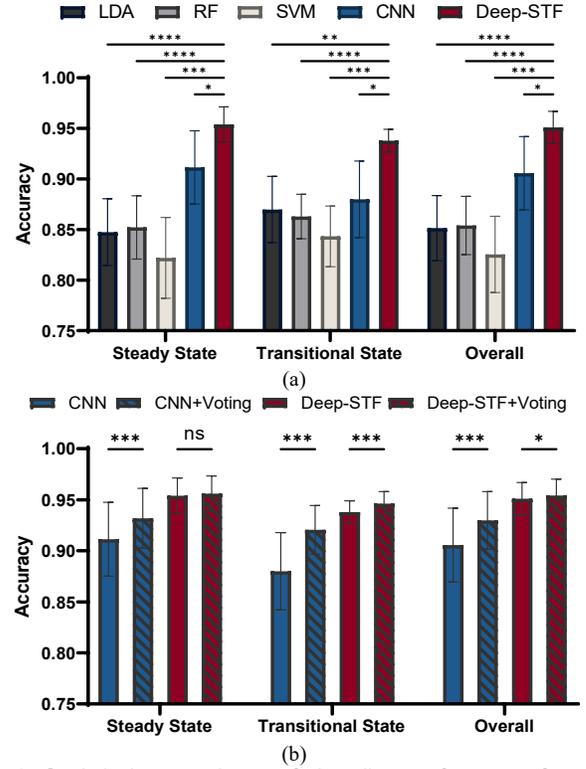

Fig. 4. Statistical comparisons of decoding performance for eight subjects using ANOVA with significance analysis: (a) comparing classification models. (b) assessing adaptive voting strategies for CNN and Deep-STF. Significance levels are denoted as follows: * p < 0.05, ** p < 0.01, *** p < 0.001, **** p < 0.0001 and "ns" denotes non-significant results within groups. Statistical comparisons across different motion conditions are not shown.

Moreover, the adaptive voting strategy effectively mitigated the decrease in prediction accuracy when used in conjunction with both CNN and Deep-STF models (p < 0.01). As prediction time increased from 100 ms to 500 ms, prediction accuracy using the CNN + adaptive voting method decreased by 5.40%, 4.84%, and 5.32% under steady-state, transition-state, and overall conditions. With the Deep-STF + adaptive voting method, prediction accuracy decreased by 3.29%, 4.15%, and 3.47% under the same conditions. The benefits of the Voting Layer strategies were more pronounced in transition-state accuracy.

### C. Prediction Effectiveness of Continuous Locomotion

The prediction effectiveness of continuous locomotion modes and transitions was evaluated in terms of stable prediction time, $P_{stable}$, and prediction rate, as depicted in Fig.6. The Deep-STF model significantly outperformed the CNN model in both stable prediction time (p < 0.01) and predict rate (p < 0.01). In Fig. 6(a), it was observed that both CNN and Deep-STF methods exhibited a certain delay compared to the labeled prediction time. As $P_{label}$ increases from 100 to 500 ms, the combination of the Deep-STF model and adaptive voting excelled, yielding a larger $P_{stable}$ ranging from 28.15 ms to 372.21 ms.

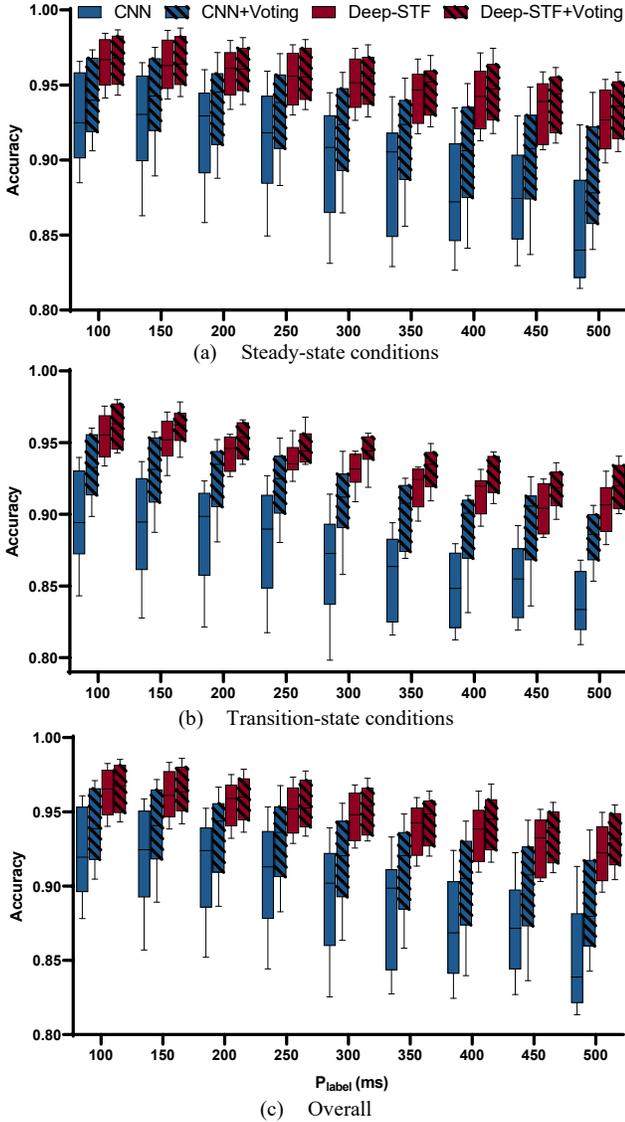

Fig. 5. Comparison of classifier configurations for intra-subject test with five subjects. (a), (b), (c) represent the averaged accuracy of steady state, transitional state, and overall samples across all prediction times.

Fig. 6 (b) presents the prediction rates corresponding to different $P_{label}$. These rates of different methods gradually increased with $P_{label}$ ranging from 100 to 250 ms, stabilizing when $P_{label}$ exceeded 250 ms. Interestingly, the combination with the adaptive voting strategy initially resulted in lower prediction rates at smaller $P_{label}$. However, this situation improved as $P_{label}$ extends, ultimately achieving comparable or even superior results after 250 ms.

### D. Channel Drop Study

A channel drop analysis was conducted to assess the effects of muscle selection on continuous locomotion prediction. The leading leg was divided into four distinct parts: the upper front leg (UF), upper back leg (UB), lower front leg (LF), and lower back leg (LB), each including three, two, one, and two major recorded muscles, respectively. As presented in Fig .7, we formed six different channel groups by combining muscles from two leg parts: UFUB, LFLB, UFLF, UBLB, UFLB, and LFUB. The "ALL" configuration utilizing all eight channels was also employed. The box plot in Fig. 7 illustrates prediction accuracy for steady state, transition state, and overall conditions within a 250 ms $P_{label}$.

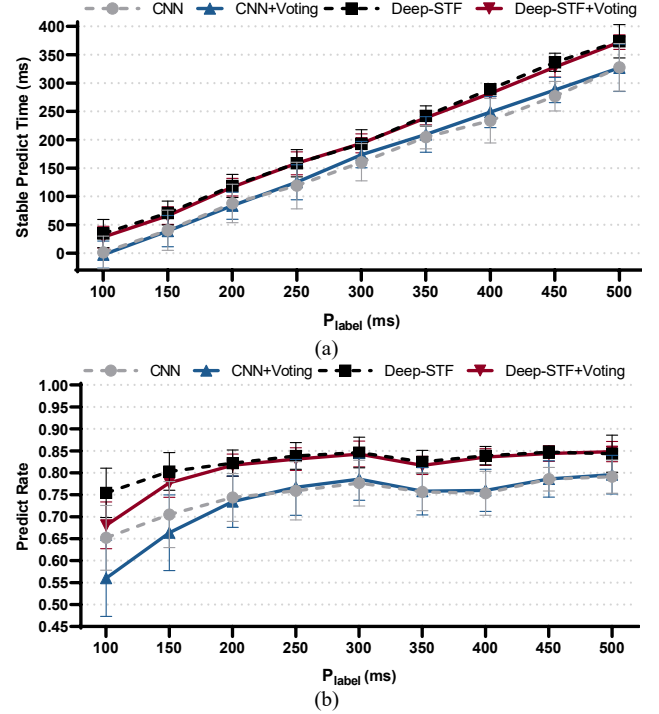

Fig. 6. (a) Illustration of stable prediction time. (b) Depiction of the observed trend in predict rates.

Among these six selected muscle groups, the UFUB, UBLB, and UFLB configurations individually achieve accuracies of 93.30%, 93.43% and 93.37% in overall prediction, with relatively minor reductions compared to the ALL configuration at 2.27%, 2.14% and 1.20% respectively. On the other hand, the LFLB and LFLB configurations, consisting of only three muscles, attain accuracies of 90.41% and 91.86% in overall prediction, reflecting more substantial reductions of 5.30% and 3.79% in comparison to the ALL configuration. Furthermore, the UFLF configuration achieves an overall accuracy of 92.90%, demonstrating a comparable reduction of approximately 2.69%.

The impact of channel-drop configurations on average $P_{stable}$ and predict rate is presented in Table IV. Reducing the number of channels results in a certain reduction in both prediction time and predict rate. Among these configurations, the UBLB configuration stands out with the smallest reduction in predict rate, ultimately achieving an average $P_{stable}$ of 128.52±29.15 ms and a predict rate of 0.81±0.04. Additionally, the channel drops in UFUB and UFLB configurations also have a relatively minor impact on prediction time, with 122.77±23.30 ms and 118.83±17.29 ms, respectively. In contrast, the LFLB, UFLF, and LFUB configurations exhibit a more substantial impact on prediction effectiveness. Among

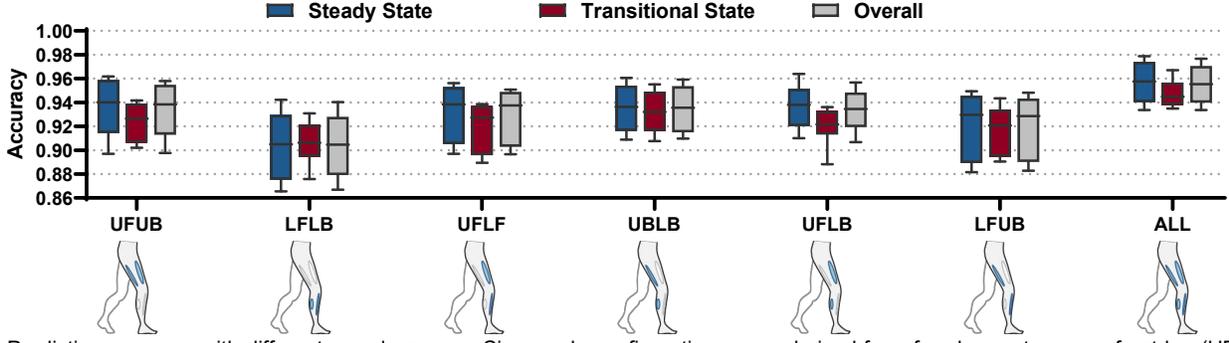

Fig. 7. Prediction accuracy with different muscle groups. Six muscle configurations were derived from four leg parts: upper front leg (UF), upper back leg (UB), lower front leg (LF), and lower back leg (LB). The accompanying box plots illustrate the accuracy performance under steady-state, transition-state and overall predictions within a 250ms prediction time for labelling.

<div align="center">

TABLE III

EFFECTS OF CHANNEL DROP CONFIGURATIONS ON STABLE PREDICTION TIME AND PREDICT RATE. ALL RESULTS ARE REPRESENTED AS MEAN ± 1 STANDARD DEVIATION ACROSS ALL SUBJECTS.

</div>

| | UFUB | LFLB | UFLF | UBLB | UFLB | LFUB | ALL |
|---|---|---|---|---|---|---|---|
| Number of muscles | 5 | 3 | 4 | 5 | 3 | 5 | 8 |
| $P_{stable}$ (ms) | 122.77±23.30 | 102.31±23.59 | 101.53±28.15 | 128.52±29.15 | 118.83±17.29 | 112.45±20.13 | 158.54±21.33 |
| Predict Rate | 0.80±0.04 | 0.76±0.04 | 0.78±0.05 | 0.81±0.04 | 0.79±0.04 | 0.79±0.04 | 0.86±0.03 |

these, the LFLB configuration experiences the largest decrease in $P_{stable}$, dropping to 101.53±28.15 ms, and a decrease in predict rate to 0.76±0.04.

## IV. DISCUSSION

Continuous prediction of locomotion intentions holds significant importance for lower-limb exoskeleton applications [38]. In this work, we present a framework to predict nine continuous locomotion modes, including 15 mode transitions between them. The experimental results demonstrated the efficacy of our proposed deep learning model, Deep-STF, in combination with the adaptive voting strategy, which effectively learned sEMG features in temporal, spatial, and frequency domains. To the best of our knowledge, this is the first exploration of the potential of sole sEMG signals in locomotion mode prediction using deep learning techniques. With a $P_{label}$ of 250 ms, our framework achieved an overall accuracy of 95.41±1.51% for predicting these continuous locomotion modes from sEMG signals. We also discovered the prediction performance concerning accuracy and prediction effectiveness across different $P_{label}$ settings. As $P_{label}$ increased from 100 ms to 500 ms at 50 ms intervals, our framework yielded a spectrum of results: overall accuracy values ranging from 93% to 96.48%, and $P_{stable}$ fluctuating between 28.15 ms and 372.21ms. In addition, the prediction accuracy of the proposed method surpassed not only traditional LDA, RF, and SVM which are highly dependent on hand-crafted features, but also the deep learning-based CNN model.

Five different methods were used for the prediction of continuous locomotion modes. LSTM [39] and CNN [40, 41] have proven to have great potential in sEMG-based human motion decoding. Their capacity for effective feature extraction via learning mechanisms surpassed the performance of traditional methods reliant on hand-crafted features. The proposed Deep-STF model was inspired by the principles of multidimensional feature learning approaches [42, 43]. This inspiration was drawn from the intricate and multidimensional nature of sEMG signals, particularly their unique representations across diverse temporal, spatial, and frequency domains. The results in Fig. 4 (a) indicate the benefit of a multidimensional feature learning model with higher accuracy than machine learning methods and a single CNN architecture (p<0.05).

To further enhance system stability and reliability, a voting strategy was employed. Muscle activities are intricately linked to both current and future gait states, introducing some fluctuations into the prediction process. Majority voting methods, as reported in previous works [40, 44, 45], have been widely adopted to bolster the stability and accuracy of sEMG-based motion detection. In our approach, we implemented an adaptive voting mechanism that combined predictions from the five adjacent sliding windows and leverages a shallow fully connected network for weight learning. Figure 4(b) underscore the efficacy of adaptive voting in enhancing the prediction accuracy of both CNN and Deep-STF methods (p < 0.05). Additionally, as revealed in Fig. 5 and Fig. 6, this mechanism demonstrated its ability to mitigate the decline in accuracy and $P_{stable}$ that accompanies an increase in $P_{label}$ from 100 ms to 500 ms.

To further illustrate the feasibility of predicting continuous locomotion modes from sEMG and other signals, as well as to assess the performance of our proposed methods, we conducted a comparative analysis of similar locomotion prediction tasks, as summarized in Table IV. In previous studies, the accuracy of locomotion intention detection relying solely on sEMG signals often fell short when compared to sensor fusion approaches. Notably, in Liu's work [25], the utilization of unilateral pure sEMG signal sensors resulted in a reduction in accuracy of over 10% for both steady and transitional-state predictions across various gait states, as compared to using sEMG in conjunction with mechanical sensors. The prediction methods involved traditional machine learning models including muscle synergies

| Author | Sensor | Methods | $N_{Locomotions}$ | $N_{Transitions}$ | Accuracy (%) | | | Prediction time (ms) |
|---|---|---|---|---|---|---|---|---|
| | | | | | Steady | Trans | Overall | |
| Y. X. Liu [25] | IMU, sEMG | Muscle synergies | 8 | 12 | \ | \ | 94.50 | 300-500 |
| B. Y. Su [37] | IMU | CNN | 5 | 8 | \ | \ | 94.15 | \ |
| D. Xu [36] | IMU, load cell | LDA | 5 | 8 | \ | \ | 93.21 | \ |
| J. Camargo [35] | IMU, sEMG, goniometer | DBN, LDA | 6 | 10 | 99.30 | 96.45 | \ | \ |
| T. Afzal [24] | sEMG | Muscle synergies | 5 | 8 | \ | \ | 82.00 | 200 |
| F. Peng [26] | sEMG, IMU | HMM | 5 | 8 | 97.80 | 91.00 | \ | 70 |
| H. Huang [27] | sEMG, load cell | SVM | 6 | 5 | \ | \ | 95.00 | 300 - 650 |
| Y. S. Wang [34] | sEMG, IMU | SVM, LDA | 5 | 8 | \ | \ | 96.00 | 300 |
| **This work** | sEMG | Deep-STF | **9** | **15** | 96.55 | 96.17 | 96.48 | 28.15 - 372.21 |

[24, 25], LDA [34, 36], SVM [27, 34], and HMM [26], as well as deep learning models including CNN [37] and DBN [35]. It is evident that our proposed Deep-STF model, operating solely on sEMG signal inputs, achieved accuracy and operation times comparable to the methods mentioned, despite the experiment scenarios we designed encompassing a broader array of locomotion modes and transitions.

In real-time exoskeleton applications, streamlining the sensor count can simplify system optimization and reduce complexity. Signal disruptions due to communication issues or sensor failures are also common. Therefore, a robust classifier capable of adapting to sudden input signal changes is essential for enhancing exoskeleton controller stability. We conducted experiments involving the removal of sEMG channels by considering muscles from two regions on the front or back sides of the upper or lower leg. Our proposed Deep-STF model maintained the overall performance even when channel data were dropped. Among the six designed channel drop configurations, the UFLB configuration, incorporating five muscles, outperformed the UFUB configuration with an impressive prediction accuracy of 93.37%. The UFLF and UBLB configurations, each with four muscles, showed good prediction efficacy, with the UFLF muscle group achieving a better prediction accuracy of 92.90%. Lastly, the LFLF and LFUB configurations, consisting of three muscles, indicated solid prediction efficacy, and the LFUB muscle group yielded a better prediction accuracy of 91.84%. Our proposed prediction framework has showcased a remarkable capability to effectively manage missing data. It has become evident that the success of locomotion prediction is influenced not only by the number of sEMG channels but also by the specific combinations of muscles in different leg regions. These findings offer valuable guidance for sensor selection and optimization in real-world exoskeleton applications.

One limitation of our study is that we only optimized and validated the model performance in an offline setting. Prior research has indicated that the use of powered exoskeletons can significantly influence users' gait patterns and muscle activation [46]. While our comprehensive offline analysis demonstrated the model's considerable potential in predicting locomotion modes on both steady and transitional states, it is necessary to conduct online validation to fully assess the model's performance when integrated into a real-world exoskeleton. Secondly, it is well known that a wide array of locomotion transitions exists in our daily life movement. In contrast, our study employed a fixed laboratory locomotion paradigm, which can be considered as the other limitation of this study. To address this, future investigations will explore more diverse real-world locomotion scenarios. Furthermore, we will investigate the model's integration with an exoskeleton to adapt to the specific requirements of exoskeletons usage. This integration will facilitate the evaluation of the model's real-time performance when actively controlling an exoskeleton, providing a more comprehensive insight into its capabilities.

## V. CONCLUSION

In this paper, we aimed to enhance locomotion modes prediction in continuous gait tasks using only sEMG signals. For that, we designed a continuous locomotion paradigm including nine distinct locomotion modes and 15 transitional states. The proposed Deep-STF model in combination with the an adaptive voting strategy were evaluated by comparing with other machine learning models and CNN on the same database. We also investigated the impact of different advance time intervals, ranging from 100 to 500 ms, on classification accuracy and stable prediction time. The experimental results revealed the following key findings: Deep-STF achieved the highest classification accuracy of 96.48% with a 100 ms advance time, accompanied by an averaged stable prediction time of 28.15 ms for transition detection. With an extended prediction time window of 500 ms, it could predict the next mode transition at stable prediction time of 372.21 ms while maintaining an average accuracy of 93%. The implementation of an adaptive voting strategy effectively mitigated the decline in classification accuracy while concurrently extended the stable prediction time as the advance time increased. These findings highlight the performance and adaptability of the Deep-STF model in continuous locomotion mode prediction.